\pdfoutput=1

\documentclass[11pt]{article}

\usepackage[]{acl}

\usepackage{times}
\usepackage{latexsym}
\usepackage{enumitem}
\usepackage{amssymb}
\usepackage{hyperref}
\usepackage[T1]{fontenc}

\usepackage[utf8]{inputenc}

\usepackage{microtype}

\usepackage{inconsolata}

\usepackage[]{todonotes}

\title{On Large Language Models' Hallucination with Regard to Known Facts}

\author{
 \textbf{Che Jiang$^{1}$\Thanks{~The work was done when Che Jiang worked as intern at Pattern Recognition Center, WeChat AI, Tencent Inc, China.}}, \textbf{Biqing Qi$^{1}$}, \textbf{Xiangyu Hong$^{1}$}, \textbf{Dayuan Fu$^{1}$}\\
 \textbf{Yang Cheng$^{1}$}, \textbf{Fandong Meng$^{2}$}, \textbf{Mo Yu$^{2\dag}$}, \textbf{Bowen Zhou$^{1}$\Thanks{~Corresponding authors}}, \textbf{Jie Zhou$^{2}$}
\\
$^1$ Department of Electronic Engineering, Tsinghua University\\
$^2$ Pattern Recognition Center, WeChat AI, Tencent Inc, China \quad 
\\
\texttt{jc23@mails.tsinghua.edu.cn} \texttt{moyumyu@global.tencent.com}\\ \texttt{zhoubowen@tsinghua.edu.cn}
\\
\\
}

\usepackage{graphicx}
\begin{document}
\maketitle
\begin{abstract}
Large language models are successful in answering factoid questions but are also prone to hallucination.
We investigate the phenomenon of LLMs possessing correct answer knowledge yet still hallucinating from the perspective of inference dynamics, an area not previously covered in studies on hallucinations.
We are able to conduct this analysis via two key ideas.
First, we identify the factual questions that query the same triplet knowledge but result in different answers. The difference between the model behaviors on the correct and incorrect outputs hence suggests the patterns when hallucinations happen.
Second, to measure the pattern, we utilize mappings from the residual streams to vocabulary space.
We reveal the different dynamics of the output token probabilities along the depths of layers between the correct and hallucinated cases. 
In hallucinated cases, the output token's information rarely demonstrates abrupt increases and consistent superiority in the later stages of the model.
Leveraging the dynamic curve as a feature, we build a classifier capable of accurately detecting hallucinatory predictions with an 88\% success rate. 
Our study shed light on understanding the reasons for LLMs' hallucinations on their known facts, and more importantly, on accurately predicting when they are hallucinating. Code and dataset are available on \url{https://github.com/dcdsf321/known_fact_hallucination.git}.

\end{abstract}

\section{Introduction}

Large Language Models (LLMs) have shown great potential to acquire extensive knowledge and apply it in various tasks \cite{petroni2019language, alkhamissi2022review, cohen2023crawling}. Despite their proficiency in generating coherent and contextually relevant text, these models frequently manifest 'factual hallucinations,' significantly undermining their reliability in practical applications \cite{zhang2023siren,huang2023survey,10.1145/3555803}. Factual hallucination is one of the least noticeable types of erroneous output, as the model often expresses fabricated content with a confident tone.

\begin{figure}[t] 
    \includegraphics[width=\columnwidth]{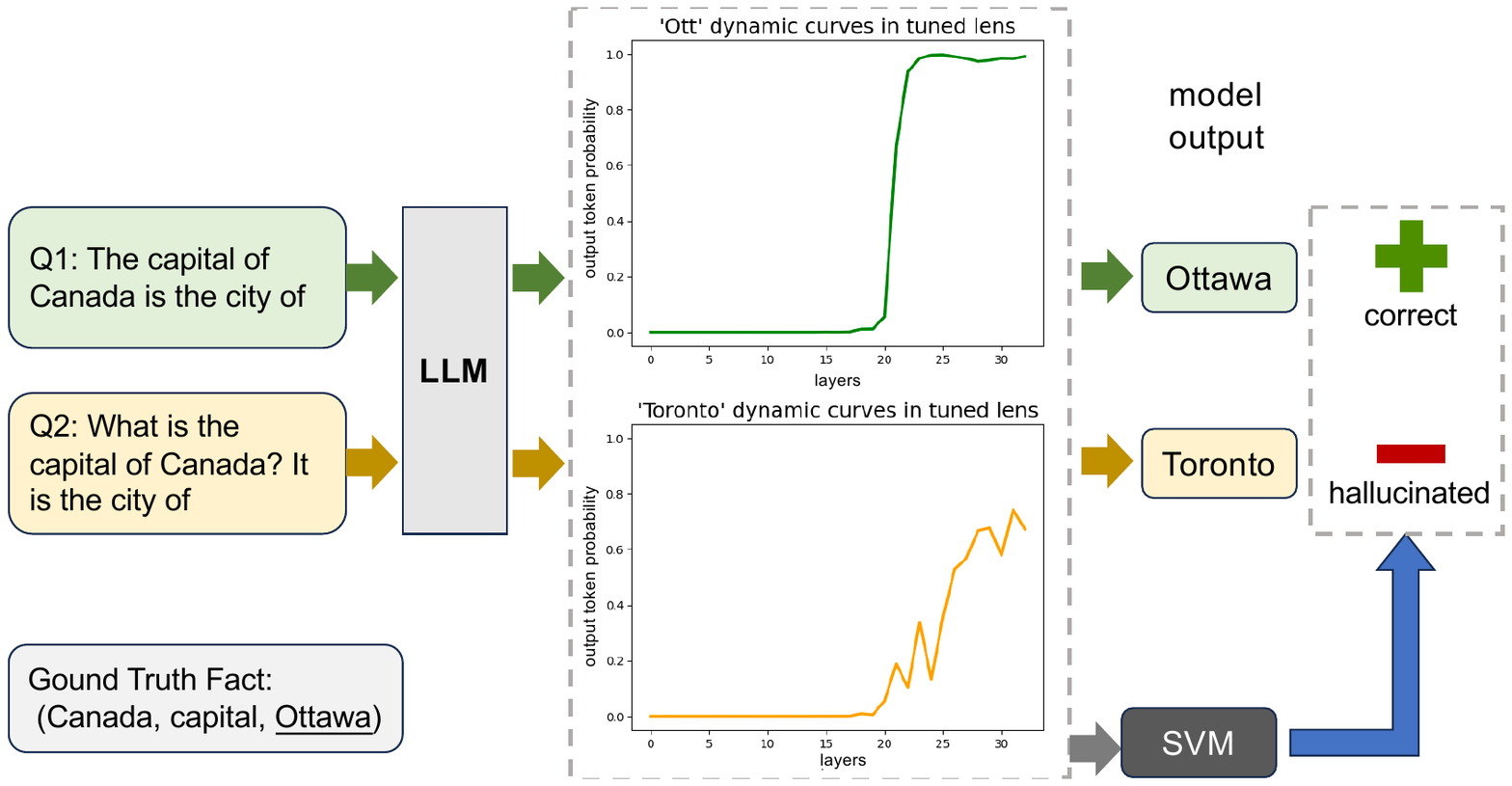} 
    \caption{We observe the difference between output token dynamics when language model makes known fact hallucinations. Using this pattern, we use a simple SVM to classify when model hallucinates.}
    \label{fig:teaser}
\end{figure}

There are two potential sources of factual hallucinations. The first arises from insufficient knowledge within the model's parameters, resulting in inaccurate responses based on a combination of false information. Consequently, the model should express uncertainty or refer to external knowledge bases when faced with such limitations. Studies propose self-assessment methods for models to rectify this form of hallucination \cite{kadavath2022language, yin-etal-2023-large, Shuster2021RetrievalAR}.
The second scenario occurs when the model's parameters memorize relevant knowledge but lack generalization ability. Prompt engineering or prefix tuning can partially alleviate these factual hallucinations, enhancing the model's performance on specific tasks \cite{Zhong2021FactualPI, Youssef2023GiveMT}. However, the mechanism behind the model's hallucination of previously memorized knowledge remains puzzling.

It is challenging to ascertain what the model does not know \cite{yin-etal-2023-large,Turpin2023LanguageMD}, but if the model provides the correct answer in response to a specific knowledge query, it can be inferred that the model has memorized relevant information. Hence, this study focuses on hallucination for known facts. Specifically, under the task of object completion for triplet knowledge, we aim to observe the behavioral characteristics of the model when there is a failure in recalling parameterized knowledge. We define a \textbf{known fact hallucination} as when a model, queried with different prompts for the same knowledge triplet, produces both correct and incorrect outputs. Incorrect outputs may include uncertain responses, irrelevant information, or incorrect entities.

In this paper, we investigate the dynamic inference characteristics of parameterized factual knowledge recall when LLM exhibits known fact hallucinations. To achieve this, we curated an information extraction dataset and filtered specific model-selected data where the accuracy varies for the same knowledge under different queries. Subsequently, we compare the inference dynamics under different scenarios so as to understand and identify such hallucinations. Our primary findings are as follows:

\setlist{nosep}
\begin{enumerate}[leftmargin=*]
    \item \textbf{Known fact hallucination arises from failed knowledge recall.}
    Our analysis shows that when model generates incorrect outputs, on average the correct answers pop to the top rank with a 30\% frequency across the layers during inference, which is significantly lower than the 78\% frequency when the output is correct.
    \item \textbf{MLP modules have a more significant impact on incorrect outputs than attention modules.} In contrast to attention modules, the Multi-Layer Perception (MLP) not only diminishes the probability of the correct answer when producing incorrect outputs but also contributes to generating erroneous outputs in the final decoding layer.
    \item \textbf{Observation of patterns in output token inference dynamics.} In the residual stream generating correct outputs, the information of the output token shows a steep increase in the middle to later layers, while erroneous outputs tend to speculate from shallower layers.
    \item \textbf{The dynamic patterns of output tokens can be used for accurate hallucination detection in predictions.} By leveraging the dynamic curve of output tokens across layers, classifiers can be trained to distinguish whether the model is recalling or hallucinating.
    We show the dynamic patterns are strong features that achieve an 88\% successful detection rate.
\end{enumerate}

\section{Related Work}

The process of knowledge recall in the model is intricate. From the perspective of grokking \cite{grok}, information with good generalization tends to occupy more condensed positions within the parameters. Prior research on locating and extracting triplet knowledge within the model indirectly supports this notion.  When performing linear mappings for relations within specific layers concerning triplet information, fewer than half of the relations achieved satisfactory results \cite{Hernandez2023LinearityOR}. These relations mainly encompass common knowledge and verbal triplets, which are high-frequency occurrences within the training data. Moreover, when pinpointing the attention head for subject-attribute mapping, only 30\% of the knowledge could be localized to a single attention head \cite{geva2023dissecting}. 
Consequently, errors occur in the model's lack of generalization regarding known knowledge, involving complex mechanisms in the reasoning process. Therefore, gaining access to model's hidden states in a broader perspective during inference process becomes imperative.

Several related studies aim to dissect internal knowledge extraction mechanisms. Causal tracing identifies influential model components during inference \cite{meng2022locating,meng2022mass}. But the addition of Gaussian noise to the input may cause artificial behavior in models, revealing a gap between causal tracing and the intricacies of natural language processing \cite{zhang2023towards}. Probing techniques employ mapping functions to detect model states or representations \cite{alain2016understanding}, connecting the model's latent space with human-understandable representations such as the truthfulness of a statement \cite{azaria2023internal, li2023inference, slobodkin2023curious}. Some work explains the model's behavior at the token level \cite{yuksekgonul2024attention,yu2019rethinking,chang2020invariant}. However, These methods do not delve into the knowledge recall dynamics. 

Regarding the exploration of internal model knowledge, numerous studies have analyzed the model from the perspective of residual streams, providing insightful breakdowns of the roles of various modules at each layer and the knowledge extraction process \cite{haviv2022understanding,dar2023analyzing,geva2022transformer,geva2023dissecting,ferrando2023explaining}. They highlight two key processes in knowledge extraction: subject enrichment and knowledge extraction. Additionally, they delve into the storage function of Multi-Layer Perception layers in knowledge and the information transmission role of attention layers. However, these studies primarily investigate mechanisms when the model successfully recalls knowledge. Our study extends their findings, examining the internal mechanisms when the model generates known fact hallucinations.

\section{Experimental Setup}

We focus on the recall process of the object in triplet knowledge $(s, r, o)$. While previous studies have partially deciphered model's successful recall process of triplet knowledge \cite{geva2023dissecting}, our curiosity lies in understanding the inference process during instances of known fact hallucination. What differences are in the dynamic change of hidden states throughout the residual stream comparing successful knowledge recalls and the failed ones?

Therefore, our experiments collect knowledge query data specifically for this scenario and test them on widely used Llama model. Then we use various tools to interpret and identify language model's dynamic inference processes when it makes known fact hallucination.

\subsection{Dataset}
We modify queries in the \textsc{CounterFact} dataset \cite{meng2022locating} by devising various ways of a query for the same relation, generating over 30k statement sentences or question-answer pairs ending with the object from triplet knowledge. The text before the object serves as the input prompt for the model, while the object itself represents the correct answer the model needs to produce. During the process of modifying statements involving triplets, we particularly focus on resolving the ambiguity in sentences after removing the object. For instance, when querying the relation called "the position of a ball-game player", a query such as "\{\textit{subject}\} plays as" might induce ambiguity, as the phrase that follows could be different from the semantic meaning "player's position." Therefore, we manually expand it to "\{\textit{subject}\}, plays in the position of" or "Which position does \{\textit{subject}\} play? \{\textit{subject}\} plays as", aiming to induce the model's understanding of the required triplet knowledge to the fullest extent. In other words, if the model continues to complete erroneous words after the modified sentence, often these words belong to the same semantic category as incorrect entities or result in irrelevant or uninformative statements, aligning with our problem settings. All the prompts are provided in Appendix.\ref{sec:appendix_data}. We followed the numbering of the triple relation categories for the \textsc{CounterFact} dataset, where each ID starting with "P" in this article represents a factual relation type.

\subsection{Model}

We use Llama2-7B-chat \cite{Touvron2023Llama2O} as the subject of analysis. It is commonly used in recent works about LLMs' hallucination \cite{yuksekgonul2024attention,chuang2023dola,li2023inference}. The instruction finetuning enhances its zero-shot ability for our task. It has a typical Transformer architecture with a model depth of $L=32$ layers, a hidden state dimension of $d=4096$, and a vocabulary size of $V=32000$. For ease of explanation and notation, we provide a brief overview of the core architecture of this model, following the annotations in \citealp{geva2023dissecting}. Assuming the input of $T$ tokens ${t_1,...,t_T}$, each token passes through an embedding matrix $E\in \mathbb{R}^{V\times d}$, transforming from the vocabulary space to the model space. Subsequently, they traverse through $L$ transformer blocks, continuously evolving within the model space, generating a residual stream of shape $T\times L\times d$. Between layer $l-1$ and $l$, the $i$-th token's hidden state $\mathbf{x}_i^{l-1}$ is updated by:
\begin{equation}
    \mathbf{x}_i^{l}=\mathbf{x}_i^{l-1}+\mathbf{a}_i^{l} + \mathbf{m}_i^{l},
\label{equ:res}
\end{equation}
where $\mathbf{a}_i^{l}$ and $\mathbf{m}_i^{l}$ are the outputs from the $l$-th attention and MLP modules. Finally, the tokens pass through an unembedding matrix $W_U^{d\times V}$, mapping back to the vocabulary space before decoding. Noting that the unembedding mapping in Llama does not have a bias term, denoted as $\mathbf{b}_U=\mathbf{0}$. 

To maintain consistency and avoid decoding strategy influence on analysis, we fix the model's decoding strategy as greedy, selecting the token with the highest probability as current output. As our constructed dataset demands the model to strive for outputting the correct answer as early as the first token, we assess the correctness of the model's output by examining whether the first 10 tokens contain the answer. The samples containing negation terms and words akin to multiple-choice answers were filtered out from the correct samples.

\subsection{Observation methods}
\label{sec:method}
\textbf{Logit Lens}, initially introduced in \cite{logitlens}, enables mapping from the model space to the vocabulary space at each position within the residual stream. This technique has been pivotal in interpreting internal representations and weight matrices of Transformer models \cite{Hanna2023HowDG, dar2023analyzing, geva2022transformer}. This allows observation of which internal positions within the model's states are already decodable to produce the final output.

\textbf{Tuned Lens}, an advancement over Logit Lens discussed in \cite{Belrose2023ElicitingLP}, acknowledges the model's inconsistent readiness for final decoding across different positions. It involves training transformations at various layers within the model space. This enhancement allows for observing changes in the internal model state, particularly in more abstract or semantic representations \cite{Halawi2023OverthinkingTT, Biderman2023PythiaAS, Nanda2023EmergentLR}.

In Section \ref{sec:lens}, we observe the transformation of the hidden state $x_T$ corresponding to the last token of the input as the number of layers increased under two different lens (methods of probability mapping for the vocabulary).The reason we only look at the last input token is that the decoding of $x_T^L$ corresponds to predicting the next token. Additionally, in practice, lens observation at positions $t<T$ concerning output tokens is minimal. For a given knowledge triplet $(s, r, o)$, we select a pair of model outputs: one considers correct $(p_r, a_r)$ and the other incorrect $(p_w, a_w)$, where $p_r$ and $p_w$ represent two different queries for the same knowledge, and $a_r$ and $a_w$ denote the output's first token. Notably, $a_r$ aligns with the first token of the ground truth object.

We observe three types of token variation curves concerning the number of layers:
\begin{itemize}[leftmargin=*]
    \item Successful recall (\textbf{Suc.}): Observing the dynamic of $a_r$ when the model input is $p_r$. This signifies the successful recall process of the relevant knowledge.
    \item Failed recall (\textbf{Fail.}): Observing the dynamic of $a_r$ when the model input is $p_w$. Here, it is important to ascertain why the model fail to recall the target knowledge.
    \item Hallucinated recall (\textbf{Hal.}): Observing the dynamic of $a_w$ when the model input is $p_w$. This analysis aims to determine where and how the incorrect output begins to manifest and eventually gets confirmed.
\end{itemize}

\textbf{Ablation.} In Section \ref{sec:module}, we conduct ablation method as a supplement to the logit lens approach for module contribution analysis. The objective of ablation is to observe changes in the output token by setting the hidden state of a specific position to zero at a particular position. This method allows tracing the output back to a specific position, highlighting its significance within the model's processing. Under the notation in Equation \ref{equ:res}, we traverse every hidden state $\mathbf{x}^l_i$ through out the residual stream and set $\mathbf{a}^{l}_t$ or $\mathbf{m}^{l}_t$ to zero, observing the resulting changes in the probability of the desired token output.

\section{Results}
\subsection{Accuracy Statistics}

Language models struggle to learn long-tail knowledge \cite{Kandpal2022LargeLM}. We roughly estimate subject popularity by examining the browsing counts of relevant entities' Wikipedia pages in the past year. Intuitively, this unpopular knowledge is also encountered during training, and as per previous studies, it can be memorized. However, the ways to query long-tail knowledge might be more limited, leading to increased instances of known fact hallucinations.

Does a subject's popularity significantly influence known fact hallucination? We manually categorize errors into uncertain responses, irrelevant information, or incorrect entities in 200 randomly sampled cases. As shown in Table \ref{tab:pop}, we found \emph{no significant correlation between these error types and the popularity of the knowledge}. Moreover, we analyze all knowledge that generates four types of queries and found that less frequently accessed knowledge is weakly correlated with more knowledge extraction errors. The result is shown in Figure \ref{fig:pop} in the appendix. This suggests the existence of an inference process unrelated to specific knowledge that contributes to these hallucinations.

\begin{table}
\centering
\begin{tabular}{c|ccc}
\hline
Popularity & Incorrect & Uncertain & Irrelevant\\
\hline
$<10^4$ & 28 & 12 & 11 \\
$10^4\sim10^5$ & 26 & 8 & 16 \\
$10^5\sim10^6$  & 27 & 9 & 16 \\
$>10^6$ & 28 & 9 & 14\\
\hline
\end{tabular}
\caption{Statistic of hallucination categories across different popularity subjects.}
\label{tab:pop}
\end{table}

\subsection{Lens Observation}
\label{sec:lens}


\begin{figure}[h] 
    \includegraphics[width=\columnwidth]{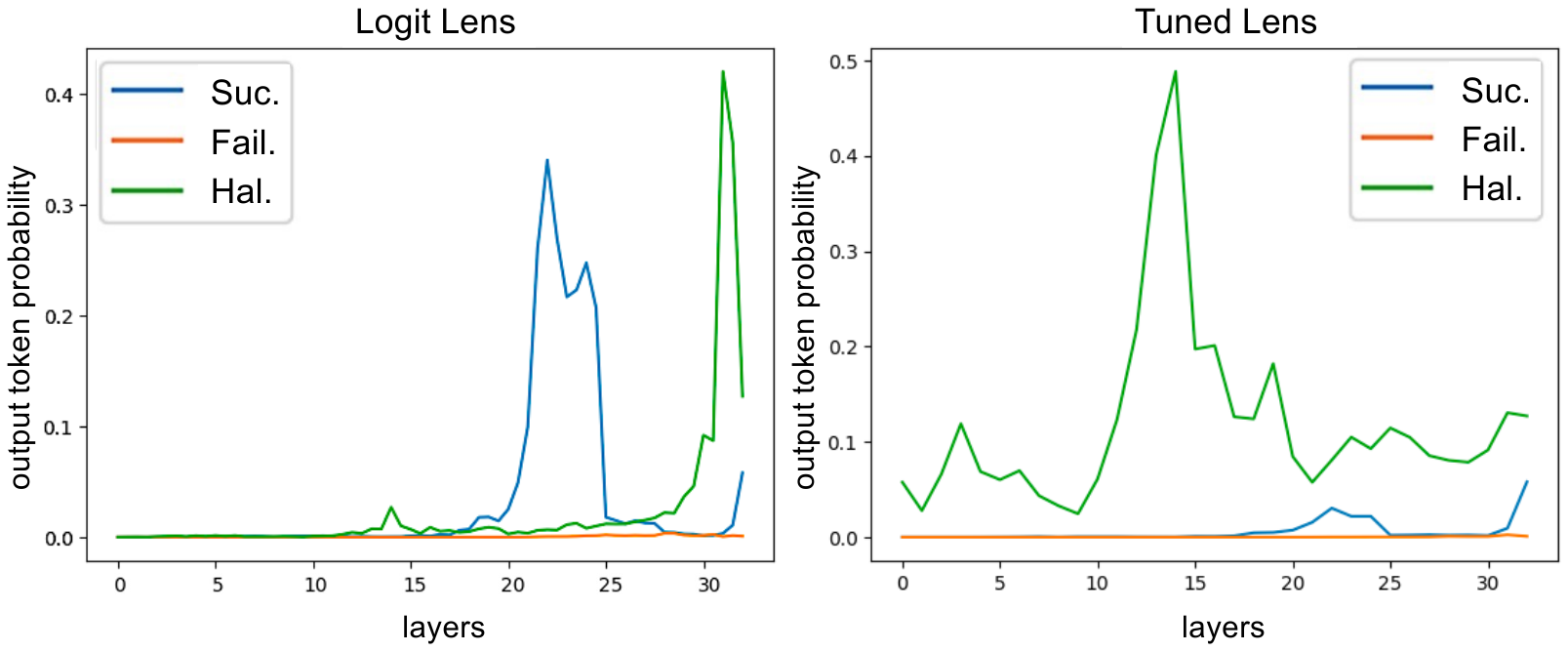} 
    \caption{An example of the variation curves in the residual stream for three types of tokens under Logit Lens and Tuned Lens. The Fail. token is not extracted at all.}
    \label{fig:lens_eg2}
\end{figure}

\begin{figure}[h] 
    \includegraphics[width=\columnwidth]{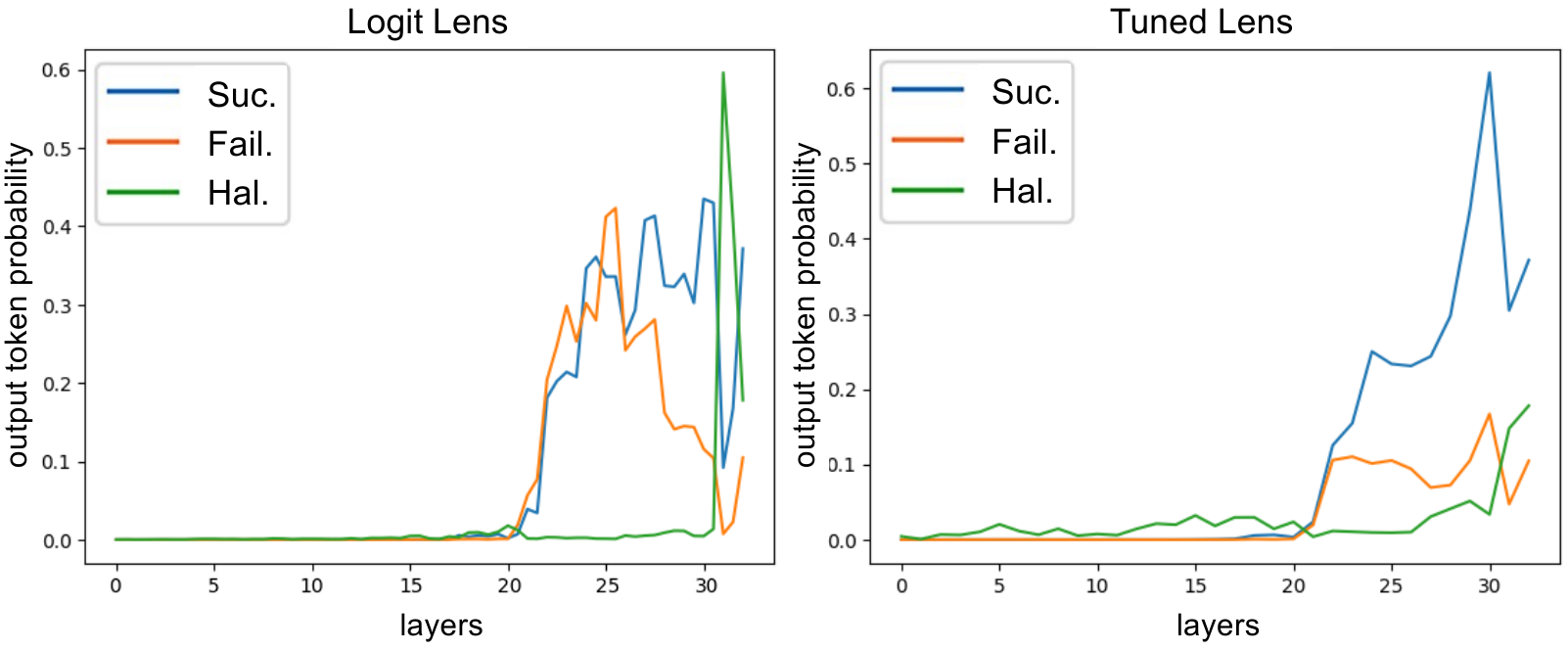} 
    \caption{An example of the variation curves in the residual stream for three types of tokens under Logit Lens and Tuned Lens. The Fail. token is temporally recalled and is suppressed afterwards.}
    \label{fig:lens_eg}
\end{figure}


\textbf{(Q1) When the model hallucinates, has the correct knowledge been retrieved?}
We demonstrate the observation of the three types of tokens discussed in Section \ref{sec:method} using Logit Lens and Tuned Lens. In Figure \ref{fig:lens_eg2}, the query's triplet is (Isaac Barrow, field of profession, mathematics). Here, $p_r$ = "The expertise of Isaac Barrow is in the field of," and $p_w$ = "What is Isaac Barrow's professional field? It is". The erroneous output is "not clear from the provided biographical information," indicating that the model failed to successfully recall the required knowledge. Correspondingly, the probability values for Fail. tokens in both Lens methods remain consistently low in the graph.

The comparison between the probability shifts for Suc. tokens and Hal. tokens in the Logit Lens reveals that the former establishes output determination earlier, whereas the latter's decoding occurs almost at the final layer. However, under observation using the Tuned Lens, it is noticeable that the model maintains conjectures about the Hal. token from the very first layer and consistently retains these assumptions throughout. In contrast, the Suc. token synchronously increases the probability of the correct token's output in line with the Logit Lens observations.

\begin{figure*}[h] 
    \centering
    \includegraphics[width=\textwidth]{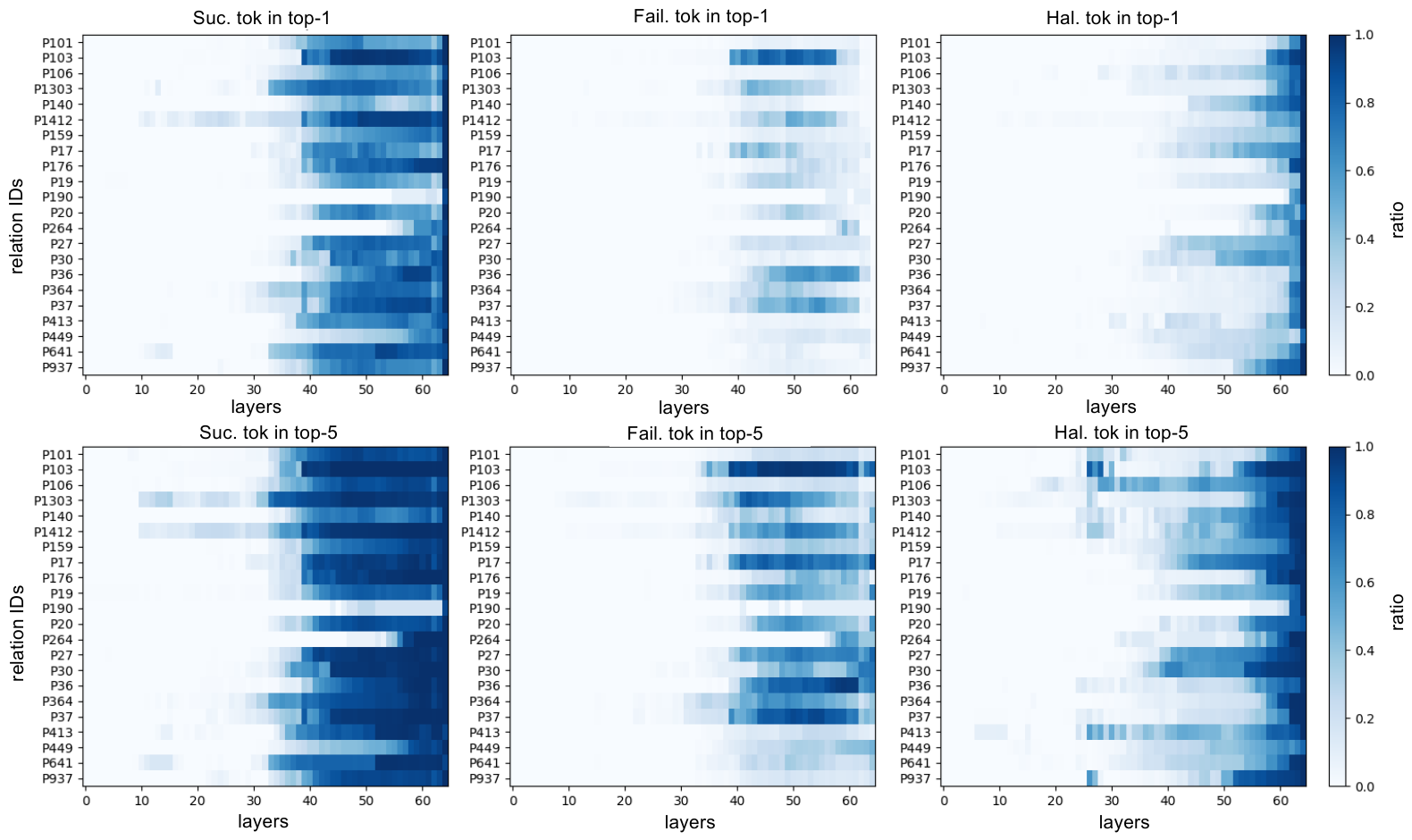} 
    \caption{The ratio of the top-1 and top-5 appearances of three types of tokens in logits rankings varies across different relations as the number of layers changes.}
    \label{fig:topk}
\end{figure*}

Does the above observation possess statistical universality? We categorize the data by relation and calculate the average probability changes of three types of tokens under two Lens observations, as shown in Figure \ref{fig:teaser}. From the results of the Tuned Lens, we observe that correctly inferred information mostly surfaces around the 20th layer, while erroneously decoded information becomes evident before the 20th layer. Regarding the Logit Lens outcomes, upon the model's confirmation of output information, there is an immediate switch to decoding mode representation, persisting through to the final layer for output. Conversely, the representation of erroneously inferred information switches to a decodable form relatively late, observed by the Logit Lens only in the last two layers. Aligned with previous research \cite{Hernandez2023LinearityOR, geva2023dissecting}, we suggest that the successful recall of knowledge indeed undergoes an 'information extraction point,' where knowledge extracted beyond a certain layer is retained and shifted to decoding mode. Erroneously decoded outputs bypass this information extraction process, being compelled to initiate decoding before being continuously conjectured up to the last layer, thus not displaying a notably high probability in the final layer's observations. From the middle column, it can be observed that in our dataset's failed knowledge recall process, the vast majority of knowledge remains unextracted.

However, a low probability observed through the lens does not imply that the token would not gain an advantage in decoding at that position. We therefore statistically analyzed the ranking of three types of tokens after Logit Lens decoding. Figure \ref{fig:topk} displays the frequency of appearance for the three types of tokens in the top 1 and top 5 at each layer for every relation. For each top-k, we first calculate the maximum occurrence frequency of each hidden state before the output layer across all relations under the Logit Lens mapping. Subsequently, we compute the average occurrence frequency for the three types of output tokens by averaging across all relations. As shown in Table \ref{tab:topk}, it is found that Fail. tokens have an average occurrence frequency of 31.28\% in the top-1 and 56.71\% in the top-5, much lower than Suc. and Hal. tokens. Hence, in most cases, the output illusion occurs because knowledge is not successfully extracted in the intermediate steps. However, it is also noticeable that some samples exhibit a phenomenon akin to Figure \ref{fig:lens_eg}, where Fail. tokens have comparable probabilities to Suc. tokens at knowledge extraction positions but get suppressed in subsequent layers, resulting in decoding failure. The Tuned Lens curve depicts the model's wavering confidence in knowledge extraction, leading to inconsistent reinforcement of correct information.

\begin{table}[!t]
\centering
\begin{tabular}{c|ccc}
\hline
  & Suc. & Fail. & Hal.\\
\hline
top1 & 77.57\% & 31.28\% & 68.04\% \\
\hline
top5 & 93.21\% & 56.71\% & 92.70\% \\
\hline
\end{tabular}
\caption{Average occurrence frequency of three kinds of tokens in top1 and top5.}
\label{tab:topk}
\end{table}

\begin{figure*}[h]
    \centering
    \includegraphics[width=\textwidth]{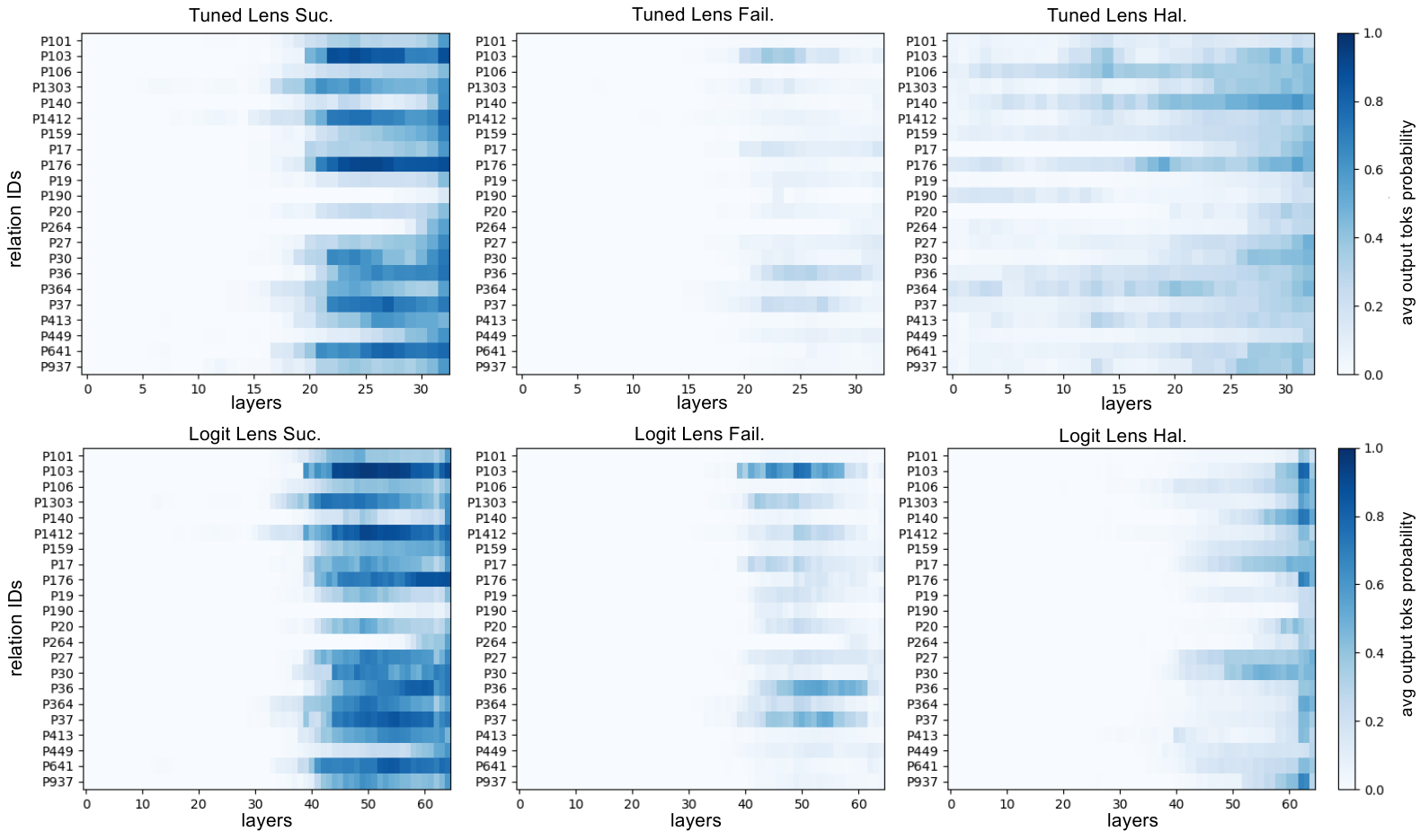} 
    \caption{Under the observation of Logit Lens and Tuned Lens, the average probability change curves of three tokens for each relation. Logit Lens has 65 values on the horizontal axis due to its output of intermediate results from the attention module.}
    \label{fig:teaser}
\end{figure*}

\subsection{Module contributions}
\label{sec:module}
\textbf{(Q2) Which module contributes more to hallucinations? What could be the potential process for this?}
From a more detailed perspective, we are interested in understanding which module contributes to errors in knowledge recall. This section of the experiment compares the roles of the MHSA (Multi-Head Self-Attention) and MLP (Multi-Layer Perceptron) modules in the success and failure of knowledge recall using two methods.

\begin{figure*}[h]
    \centering
    \includegraphics[width=\textwidth]{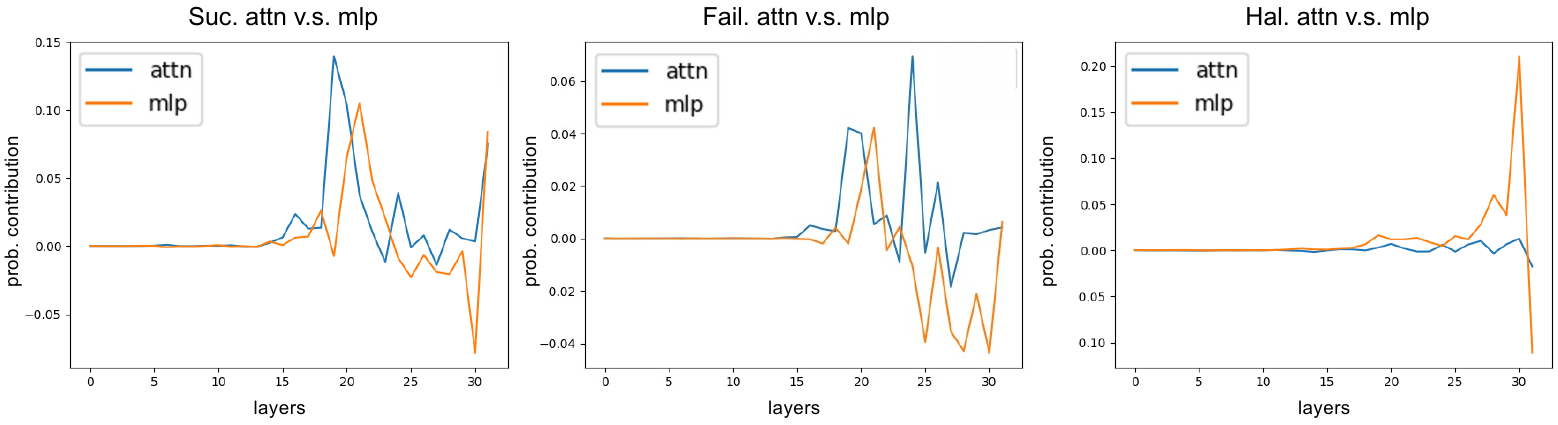} 
    \caption{The average contributions of the attention module and the MLP module to the residual stream variations of three types of tokens.}
    \label{fig:attnmlp}
\end{figure*}

The first method, inspired by the Logit Lens approach, projects the directional changes of each layer's modules on the hidden state of the last token towards the decoding matrix for the token of interest. This allows us to observe the contribution of each module at every position to the output of the three token types mentioned earlier. The results are depicted in Figure \ref{fig:attnmlp}. For successfully recalled samples, both MHSA and MLP demonstrate equally significant contributions to knowledge extraction, especially around the 20th layer, where a substantial amount of knowledge is extracted. However, for failed recalls, while some knowledge is extracted around the 20th layer, the MLP exerts a stronger inhibitory effect towards the end of the model, particularly contributing significantly to erroneous output decoding.

\begin{figure*}[h]
    \centering
    \includegraphics[width=\textwidth]{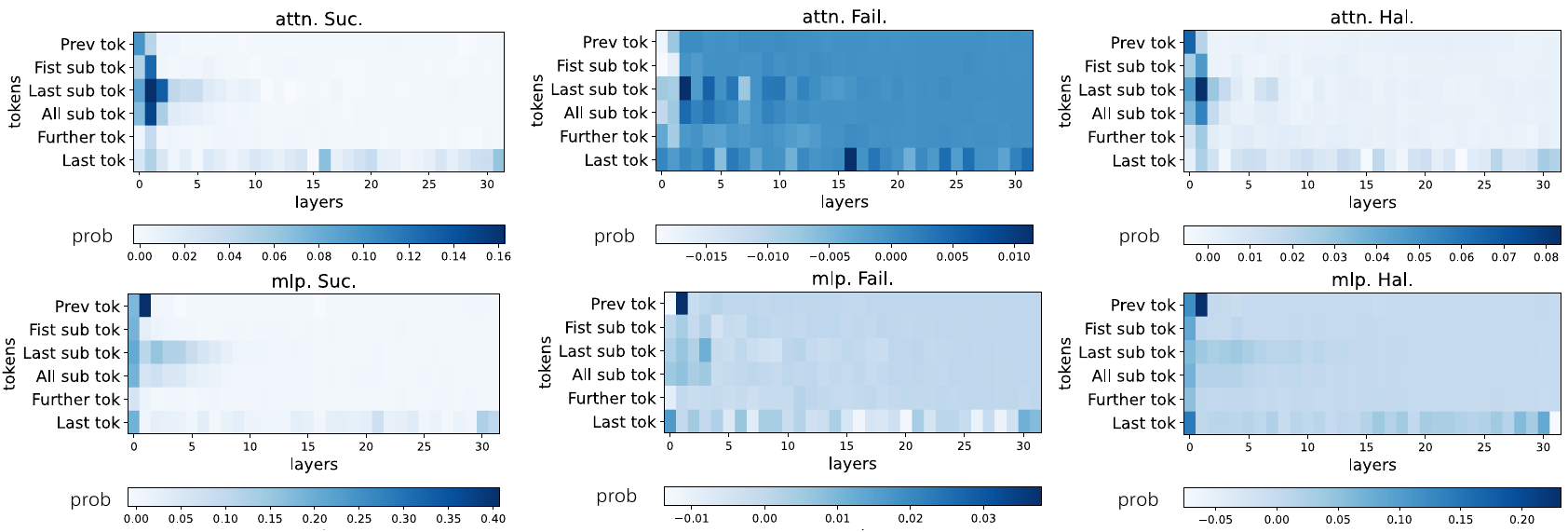} 
    \caption{The ablation results of MHSA and MLP module of three types of tokens. The darker colors in the heatmap indicate a higher positive effect on the final output.}
    \label{fig:ablation}
\end{figure*}

The second method involves modules' ablation discussed in Section \ref{sec:method}. 
The results are illustrated in Figure \ref{fig:ablation}. We sampled over 200 pairs of $(p_r, a_r)$ and $(p_w, a_w)$ from our dataset and categorized them based on token positions. In the initial half of the model, the semantic parsing of the query plays a crucial role. In comparison to successful knowledge recall, failed recalls show minimal impact on the final output from the semantic parsing results at the subject position. Following successful semantic parsing, the processing of output information mostly occurs at the position of the last token. This reaffirms the significance of considering the last token as a metric for subsequent analysis.

These two results infer internal processes within the model during knowledge recall errors. During the early stages of the model, deviations in semantic parsing of the query may lead to ineffective or insufficient extraction of internal knowledge in intermediate layers of the model. When competing with hallucinated outputs under the MLP's influence, correct tokens progressively lose prominence to illusory information, leading to their failure to compete in the final decoding stage.

\subsection{Logit evolution pattern}
The process of failed knowledge extraction is more evidently reflected in the variation of the hidden state of the last token across layers. Consequently, we propose a straightforward method to detect generations of factual hallucination by observing these specific features.

\begin{figure}[h] 
    \includegraphics[width=0.8\columnwidth]{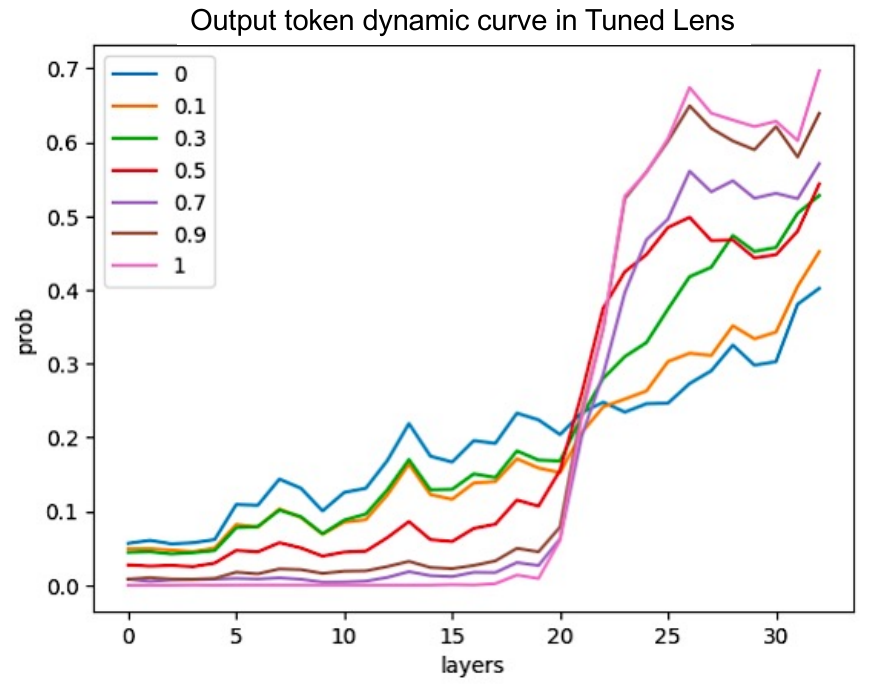} 
    \caption{The average dynamic curve of output token under Tuned Lens mapping across various correct rate ratios for relation P36.}
    \label{fig:P36}
\end{figure}

\textbf{(Q3) Are there any patterns in the inference dynamics of hallucination versus correct predictions?}
To facilitate a more intuitive comparison of alterations in output token representations, we blend successful and failed samples in varying proportions, observing the resultant probability curves using Tuned Lens mapping. Figure \ref{fig:P36} demonstrates the results for the relation 'country's capital'. When the model successfully extracts information, the probability of the output token after mapping predominantly shows a significant increase in the mid-to-late layers, with probabilities starting at 0 in the early stages. This aligns with the process of 'factual recall,' where early stages focus on query parsing and later stages on answer extraction and decoding. However, hallucination outputs do not exhibit notable leaps at relevant positions; they often contain representations of the output token before semantic parsing completes. This suggests that these tokens are likely hallucinations or incorrect answers.

\textbf{(Q4) Can we benefit from the observed patterns for automatic hallucination detection?}
Utilizing these observations, we train a linear SVM model using the probability variation curves after mapping with the two type of Lens, each is of the same length as the model layers. This model can be employed in knowledge extraction scenarios for white-box LLMs. It doesn't require knowledge of what the correct answer is; rather, it only needs to backtrack the mapping pattern of the first token output to the corresponding residual stream to determine whether the model has generated an illusionary output.

\begin{table}[h]
\centering
\begin{tabular}{c|ccc}
\hline
\textbf{Model} & \textbf{Logit} & \textbf{Tuned} & \textbf{Both}\\
\hline
Llama-7B-chat & 0.839 & 0.854 & \textbf{0.879} \\
Llama-13B-chat & 0.849 & 0.840 & \textbf{0.878} \\
OPT-6.7B & 0.856 & 0.858 & \textbf{0.865} \\
Pythia-6.9B & \textbf{0.824} & 0.764 & 0.822 \\
\hline
\end{tabular}
\caption{Hallucination classification accuracy using output token dynamics across different models.}
\label{tab:SVM}
\end{table}

Our SVM model was implemented using the SVC class (C-Support Vector Classification) from the \textit{sklearn} library. We use the default hyperparameters of the class. We will investigate the hyperparameter in future work. For all the probability variation curves obtained through the Lens methods, we performed shuffling and used 20\% of the vectors as test data, with the remaining vectors serving as training data. To validate the consistency of our observations across different models, we also conducted experiments on the Llama2-13B-chat \cite{Touvron2023Llama2O}, OPT-6.7B \cite{zhang2022opt}, and Pythia-6.9B \cite{biderman2023pythia} models. We compared three curves as vectors for SVM classification: the probability variation curves obtained using Logit Lens, the curves obtained using Tuned Lens, and the curves obtained by concatenating the two.

The results are presented in Table \ref{tab:SVM}.  Using Output token dynamics, predicting whether the four models are hallucinating has achieved an accuracy of over 80\%. Among them, the results of classification using two curves often perform better, indicating that the hallucination exhibit different patterns under the observation of different lenses. In general, output token dynamics can be used to predict the hallucinatory behavior of models regarding factual knowledge.

\section{Conclusion}

We have analyzed the scenario where LLMs hallucinate on known facts. Using our dataset based on triplet knowledge, we make comparative observations of the model's reasoning dynamics across various outputs. We show that the cause of hallucinations lies in the failure of factual recall. The failure may result from a bias in subject parsing, leading to inadequate extraction of object-related information at higher levels. This information then competes with hallucinatory information flow and gets suppressed by MLPs. Based on the distinct differences observed in the dynamics of reasoning, we train a well-performing classifier to determine the presence of hallucinations in model outputs. Leveraging the findings of this study, future work could explore the impact of query formulation on known knowledge recall and methods to mitigate such hallucinations. Our discoveries offer a novel perspective on observing knowledge hallucination: viewing the reasoning process of language models as a dynamic system where the internal state variations influence the ultimate output. This dynamic can be observed to analyze and determine the nature of the output. Subsequent research could generalize this perspective to broader forms of knowledge outputs and more complex reasoning tasks.

\section{Limitations}
While our investigation has shed light on understanding and identifying known fact hallucinations within LLMs, several limitations warrant acknowledgment. Firstly, to guarantee fair comparison for factual recall, our analysis relies on triplet knowledge datasets, potentially limiting the generalizability of our findings to other types of knowledge structures or domains. The inference dynamics observed in subject parsing and information extraction might differ concerning alternate data representations, necessitating further exploration. Additionally, our study primarily focused on a widely used transformer model, llama2-7b-chat. Future research should encompass analyses across a broader spectrum of open-source LLMs to validate and extend our findings. Furthermore, while we emphasize the dynamic nature of language model reasoning, our study offers a preliminary understanding. Comprehensive elucidation of the intricate internal state variations and their direct influence on output remains a complex area that demands deeper investigation.

\section*{Acknowledgements}
This work is supported by the National Science and Technology Major Project (No. 2022ZD0117903).
We extend our gratitude to the anonymous reviewers for their insightful feedback. We thank Kaiyan Zhang, Ermo Hua, Zixu Hao and Yiyao Jiang for their helpful comments.
\bibliography{acl_latex}

\appendix

\section{Dataset Statistics}
\label{sec:appendix_data}

\begin{figure}[h] 
    \includegraphics[width=\columnwidth]{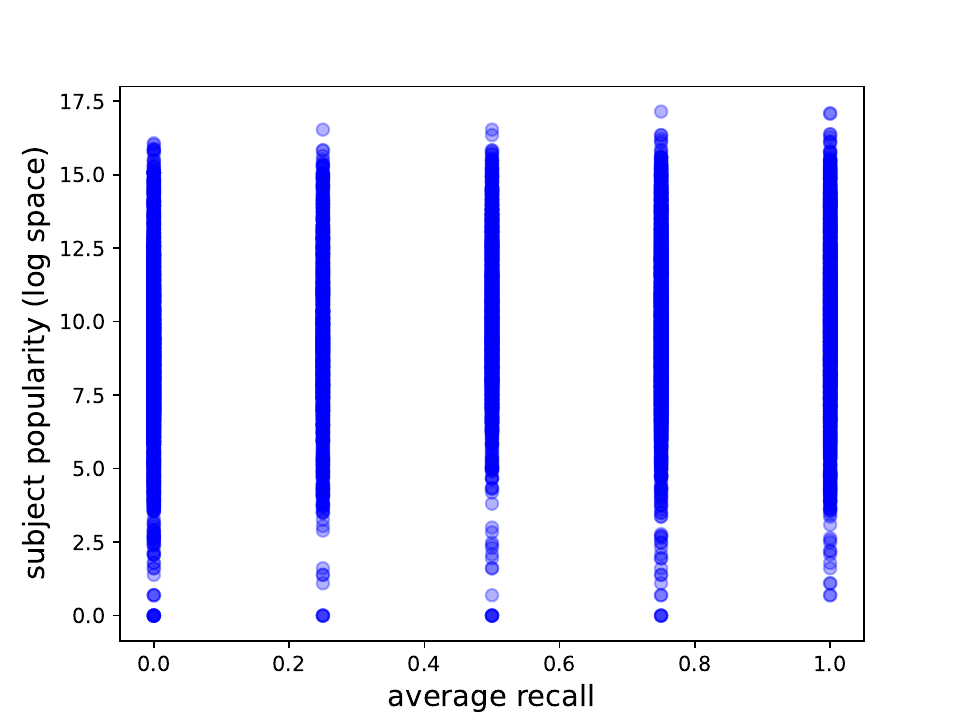} 
    \caption{Average answer recall for questions generated by subjects of different popularity.}
    \label{fig:pop}
\end{figure}

Figure \ref{fig:pop} shows the average answer recall across the four types of queries generated by subjects of different popularity. There is no significant relation between subject popularity and the result of our queries. Table \ref{relation_id A} and \ref{relation_id B} present all the the relation IDs and their corresponding relation meanings that we use in our datasets. For each type of relation, we curate four query templates to generate knowledge prompts.

\begin{table*}
\centering
\begin{tabular}{l|l|l}
\hline
\textbf{relation id} & \textbf{queries} & \textbf{prompts}\\
\hline
P17 & country of cities & \begin{tabular}{@{}p{10cm}@{}}
"\{\textit{subject}\}, which is located in the country of" \\
"\{\textit{subject}\} is located in the country of" \\
"\{\textit{subject}\} is situated in the country of"
\end{tabular} \\
\hline
P19 & city of birth & \begin{tabular}{@{}p{10cm}@{}}
"\{\textit{subject}\}, was born in the city of" \\
"\{\textit{subject}\} is originally from the city of" \\
"\{\textit{subject}\} is native to the city of" \\
"Which city was \{\textit{subject}\} born in? \{\textit{subject}\} was born in"
\end{tabular} \\
\hline
P20 & city of death & \begin{tabular}{@{}p{10cm}@{}}
"\{\textit{subject}\} expired at the city of" \\
"\{\textit{subject}\} passed away at the city of" \\
"In which city did \{\textit{subject}\}'s life end? In" \\
"\{\textit{subject}\} died in the city of"
\end{tabular} \\
\hline
P27 & country of citizenship & \begin{tabular}{@{}p{10cm}@{}}
"\{\textit{subject}\} is a citizen of the country of" \\
"To which nation does \{\textit{subject}\} belong? \{\textit{subject}\} belongs to" \\
"Which country is \{\textit{subject}\} from? \{\textit{subject}\} is from" \\
"The country that \{\textit{subject}\} belongs to is"
\end{tabular} \\
\hline
P30 & continent & \begin{tabular}{@{}p{10cm}@{}}
"\{\textit{subject}\} belongs to the continent of" \\
"\{\textit{subject}\} is located in the continent of" \\
"\{\textit{subject}\}, in the continent called" \\
"To which continent does \{\textit{subject}\} belong? \{\textit{subject}\} belongs to"
\end{tabular} \\
\hline
P36 & capital & \begin{tabular}{@{}p{10cm}@{}}
"\{\textit{subject}\}'s capital is" \\
"The capital city of \{\textit{subject}\} is" \\
"\{\textit{subject}\}, which has the capital city" \\
"What is \{\textit{subject}\}'s capital city? It is"
\end{tabular} \\
\hline
P37 & official language & \begin{tabular}{@{}p{10cm}@{}}
"In \{\textit{subject}\}, an official language is" \\
"The official language of \{\textit{subject}\} is" \\
"What language is officially used in \{\textit{subject}\}? It is" \\
"What is the official language of \{\textit{subject}\}? It is"
\end{tabular} \\
\hline
P101 & field of work & \begin{tabular}{@{}p{10cm}@{}}
"\{\textit{subject}\}'s domain of work is" \\
"The expertise of \{\textit{subject}\} is" \\
"What is \{\textit{subject}\}'s professional field? It is" \\
"\{\textit{subject}\} works in the field of"
\end{tabular} \\
\hline
P103 & native language & \begin{tabular}{@{}p{10cm}@{}}
"The mother tongue of \{\textit{subject}\} is" \\
"\{\textit{subject}\} is a native speaker of" \\
"What is \{\textit{subject}\}'s native language? It is" \\
"\{\textit{subject}\}'s native language is"
\end{tabular} \\
\hline
P106 & occupation & \begin{tabular}{@{}p{10cm}@{}}
"\{\textit{subject}\}, who works as" \\
"The profession of \{\textit{subject}\} is" \\
"\{\textit{subject}\}'s occupation is" \\
"What is \{\textit{subject}\}'s profession? It is"
\end{tabular} \\
\hline
P108 & employer & \begin{tabular}{@{}p{10cm}@{}}
"\{\textit{subject}\}, who is employed by the company called" \\
"Which company is \{\textit{subject}\} employed at? It is" \\
"The company that hired \{\textit{subject}\} is"
\end{tabular} \\
\hline
P140 & religion & \begin{tabular}{@{}p{10cm}@{}}
"What is the official religion of \{\textit{subject}\}? It is" \\
"The official religion of \{\textit{subject}\} is" \\
"\{\textit{subject}\}, a follower of the religion" \\
"\{\textit{subject}\} follows the religion of"
\end{tabular} \\
\hline
P159 & headquarters location & \begin{tabular}{@{}p{10cm}@{}}
"The headquarters of \{\textit{subject}\} is located in the city" \\
"\{\textit{subject}\}, whose headquarters is in the city of" \\
"\{\textit{subject}\} is based in the city of" \\
"Which city is \{\textit{subject}\} based in? It is"
\end{tabular} \\
\hline
\end{tabular}\\
\caption{ The table displays the relation IDs and their corresponding relation meanings, along with the prompts created for each relation.
}
\label{relation_id A}
\end{table*}

\begin{table*}
\centering
\begin{tabular}{l|l|l}
\hline
\textbf{relation id} & \textbf{queries} & \textbf{prompts}\\
\hline
P176 & creator & \begin{tabular}{@{}p{10cm}@{}}
"\{\textit{subject}\} was produced by the company called" \\
"The company produced \{\textit{subject}\} is" \\
"Which company produced \{\textit{subject}\}? It is" \\
"\{\textit{subject}\}, produced by the company called"
\end{tabular} \\
\hline
P178 & developer & \begin{tabular}{@{}p{10cm}@{}}
"\{\textit{subject}\} was developed by the company called" \\
"The company developed \{\textit{subject}\} is" \\
"Which company developed \{\textit{subject}\}? It is" \\
"\{\textit{subject}\}, developed by the company called"
\end{tabular} \\
\hline
P190 & twin city & \begin{tabular}{@{}p{10cm}@{}}
"What is the twin city of \{\textit{subject}\}? It is" \\
"The twin city of \{\textit{subject}\} is" \\
"\{\textit{subject}\} is a twin city of"
\end{tabular} \\
\hline
P264 & record label & \begin{tabular}{@{}p{10cm}@{}}
"\{\textit{subject}\}'s record label is" \\
"What is the record label for \{\textit{subject}\}? It is" \\
"\{\textit{subject}\}'s music is released by music label called" \\
"\{\textit{subject}\} is affiliated with record label called"
\end{tabular} \\
\hline
P364 & original language & \begin{tabular}{@{}p{10cm}@{}}
"The original language of \{\textit{subject}\} is" \\
"What's the original language of \{\textit{subject}\}? It is" \\
"\{\textit{subject}\} was originally filmed in the language of"
\end{tabular} \\
\hline
P407 & writing language & \begin{tabular}{@{}p{10cm}@{}}
"\{\textit{subject}\} is written in the language of" \\
"The original language of \{\textit{subject}\} is" \\
"\{\textit{subject}\}, written in the language of"
\end{tabular} \\
\hline
P413 & position played & \begin{tabular}{@{}p{10cm}@{}}
"\{\textit{subject}\} plays in the position of" \\
"Which position does \{\textit{subject}\} play? It is" \\
"\{\textit{subject}\}, who plays the position called"
\end{tabular} \\
\hline
P449 & premiere & \begin{tabular}{@{}p{10cm}@{}}
"\{\textit{subject}\} was released on" \\
"\{\textit{subject}\} premiered on" \\
"\{\textit{subject}\} was originally aired on" \\
"\{\textit{subject}\} debuted on"
\end{tabular} \\
\hline
P641 & sport played & \begin{tabular}{@{}p{10cm}@{}}
"\{\textit{subject}\} professionally plays the sport" \\
"\{\textit{subject}\} plays" \\
"\{\textit{subject}\} is a professional" \\
"What sport does \{\textit{subject}\} play? \{\textit{subject}\} plays"
\end{tabular} \\
\hline
P937 & workplace location & \begin{tabular}{@{}p{10cm}@{}}
"\{\textit{subject}\} used to work in the city of" \\
"\{\textit{subject}\} mainly worked in the city of" \\
"Which city did \{\textit{subject}\} work in? It is"
\end{tabular} \\
\hline
P1303 & instrument played & \begin{tabular}{@{}p{10cm}@{}}
"\{\textit{subject}\} plays the instrument called" \\
"\{\textit{subject}\} is skilled at playing the" \\
"Which instrument does \{\textit{subject}\} mainly play? It is" \\
"The primary instrument \{\textit{subject}\} performs on is the"
\end{tabular} \\
\hline
P1412 & spoken language & \begin{tabular}{@{}p{10cm}@{}}
"What language does \{\textit{subject}\} speak? \{\textit{subject}\} speaks" \\
"What is \{\textit{subject}\}'s primary language? It is"\\
"The language used by \{\textit{subject}\} is"\\
"The language that \{\textit{subject}\} is fluent in is" \\
\end{tabular}\\
\hline
\end{tabular}\\
\caption{
The table displays the relation IDs and their corresponding relation meanings, along with the prompts created for each relation.
}
\label{relation_id B}
\end{table*}

\end{document}